%% file: main.tex
\documentclass[runningheads]{llncs}

\usepackage[mobile]{eccv}

\usepackage{eccvabbrv}

\usepackage{graphicx}
\usepackage{booktabs}

\usepackage[accsupp]{axessibility}  

\usepackage{amsmath}
\usepackage{amssymb}
\usepackage{caption}
\usepackage{multirow}
\usepackage{url}
\usepackage{xcolor}

\usepackage[pagebackref,breaklinks,colorlinks]{hyperref}

\usepackage{orcidlink}

\begin{document}

\title{Interactive Text-to-texture Synthesis via Unified Depth-aware Inpainting}

\titlerunning{InTeX}

\author{
Jiaxiang Tang\inst{1}\thanks{Work done while visiting S-Lab, Nanyang Technological University.} \and
Ruijie Lu\inst{1} \and
Xiaokang Chen\inst{1} \and 
Xiang Wen\inst{2,3} \and
Gang Zeng\inst{1} \and
Ziwei Liu\inst{4}
}

\authorrunning{Tang et al.}

\institute{
National Key Lab of General AI, Peking University \and 
Zhejiang University \and
Skywork AI \and
S-Lab, Nanyang Technological University
}

\maketitle

{
    \vspace{-0.5cm}
    \begin{center}
    \textbf{\url{https://me.kiui.moe/intex}}
    \end{center}
    \vspace{-0.2cm}
    \centering
    \captionsetup{type=figure}
    \includegraphics[width=\textwidth]{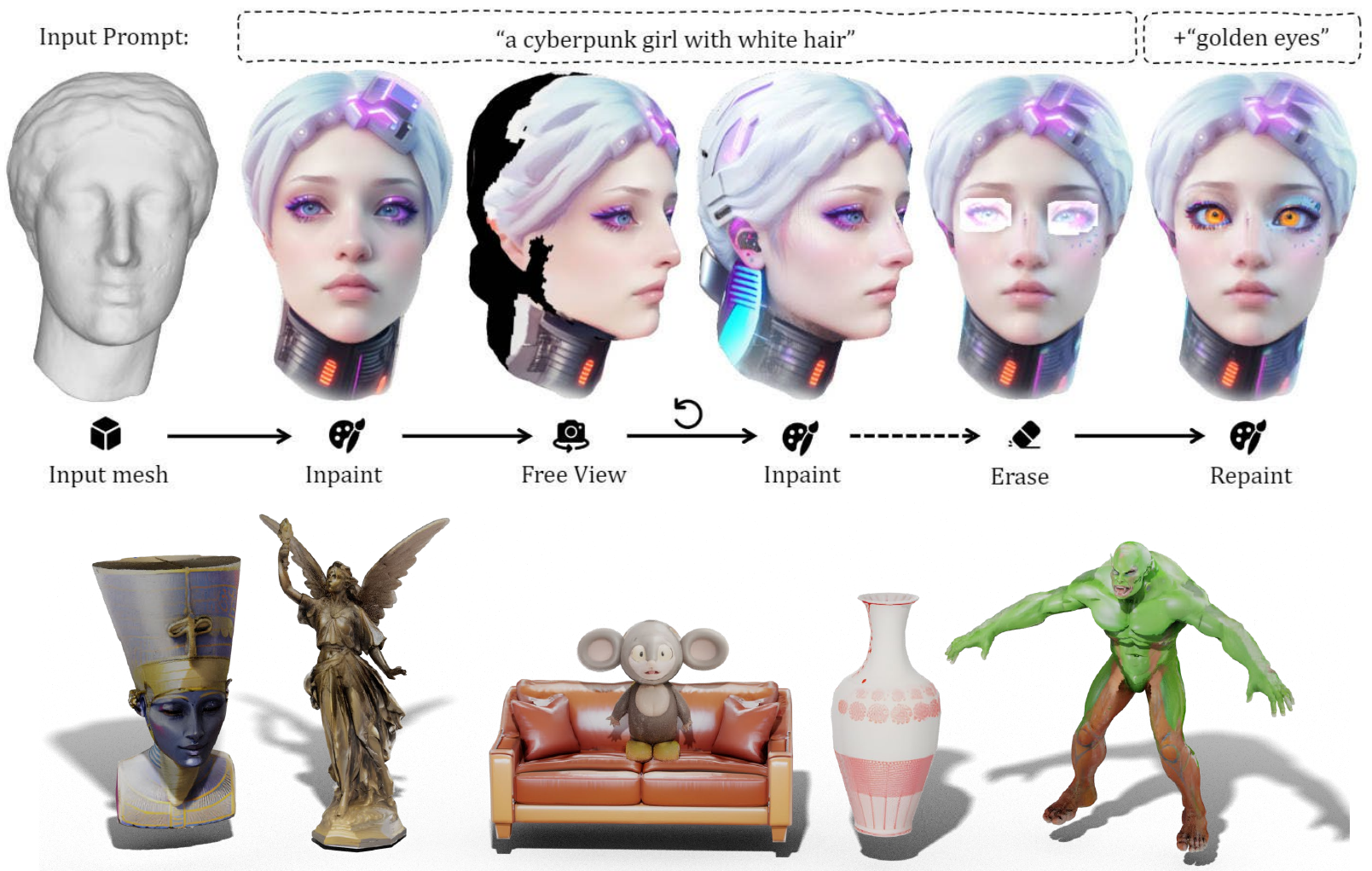}
    \captionof{figure}{
    We propose \textit{InteX}, an interactive text-to-texture synthesis framework via unified depth-aware inpainting. 
    Our method supports flexible visualization, inpainting, erasing, and repainting with a graphic user interface.
    }
    \label{fig:teaser}
}

\begin{abstract}
\input{0_abstract}
\end{abstract}

\section{Introduction}
\label{sec:intro}
\input{1_intro}

\section{Related Work}
\label{sec:rel}
\input{2_related}

\section{Methodology}
\label{sec:method}
\input{3_method}

\section{Experiments}
\label{sec:exp}
\input{4_exps}

\section{Conclusion}
\input{5_conclusion}

\subsubsection{Acknowledgements.}
This work is supported by the Sichuan Science and Technology Program (2023YFSY0008), National Natural Science Foundation of China (61632003, 61375022, 61403005), Grant SCITLAB-20017 of Intelligent Terminal Key Laboratory of SiChuan Province, Beijing Advanced Innovation Center for Intelligent Robots and Systems (2018IRS11), and PEK-SenseTime Joint Laboratory of Machine Vision. 
This study is also supported by the Ministry of Education, Singapore, under its MOE AcRF Tier 2 (MOE-T2EP20221- 0012), NTU NAP, and under the RIE2020 Industry Alignment Fund – Industry Collaboration Projects (IAF-ICP) Funding Initiative, as well as cash and in-kind contribution from the industry partner(s).

%
%
\bibliographystyle{eccv_refstyle}
\bibliography{ref}

\clearpage


\end{document}

%% file: 0_abstract.tex
Text-to-texture synthesis has become a new frontier in 3D content creation thanks to the recent advances in text-to-image models. 
Existing methods primarily adopt a combination of pretrained depth-aware diffusion and inpainting models, yet they exhibit shortcomings such as 3D inconsistency and limited controllability. 
To address these challenges, we introduce \textbf{InteX}, a novel framework for interactive text-to-texture synthesis. 
\textbf{1)} InteX includes a user-friendly interface that facilitates interaction and control throughout the synthesis process, enabling region-specific repainting and precise texture editing.
\textbf{2)} Additionally, we develop a unified depth-aware inpainting model that integrates depth information with inpainting cues, effectively mitigating 3D inconsistencies and improving generation speed. 
Through extensive experiments, our framework has proven to be both practical and effective in text-to-texture synthesis, paving the way for high-quality 3D content creation.

\keywords{3D Generation \and Texture Synthesis}

%% file: 1_intro.tex

Automatic 3D content generation stands out as a promising and widely investigated research domain, with significant implications for diverse fields such as 3D modeling, video game development, and virtual reality. An essential aspect of crafting 3D content involves the creation of high-quality textures, traditionally a labor-intensive task requiring skilled artists' manual input.
Text-to-texture synthesis addresses this challenge by aiming to automatically produce realistic and diverse textures on 3D objects guided by textual descriptions. Recent advances in denoising diffusion models~\cite{rombach2022high,meng2021sdedit,ho2020denoising} have propelled notable progress in text-to-image synthesis tasks. These methods enables rapid generation of highly realistic and high-resolution 2D images given textual prompts. Moreover, the incorporation of additional information, such as depth and normal maps, has been demonstrated to control the generation process~\cite{zhang2023adding,mou2023t2i}.
Nevertheless, text-to-texture synthesis introduces a more complex challenge. It involves the generation of consistent content on the surfaces of 3D objects, and straightforward attempts to leverage text-to-image models often fall short in producing textures of the desired quality.

Recent investigations into text-to-texture synthesis can be broadly categorized into two main lines. 
Some methods like~\cite{yu2023texture} advocate for the direct training of 3D diffusion models using 3D datasets. 
This approach aims to minimize the necessity for multiple diffusion processes when generating textures on 3D shapes. 
However, scaling up 3D diffusion models to accommodate higher resolutions and larger datasets presents substantial challenges. 
Consequently, the textures generated by these methods often exhibit blurriness and limited diversity.
In contrast, alternative techniques~\cite{richardson2023texture,chen2023text2tex,cao2023texfusion} opt to leverage pretrained 2D diffusion models and employ an iterative painting approach to generate textures. 
To enable these 2D diffusion models to comprehend 3D information, these methodologies rely on a combination of depth-to-image and image inpainting techniques during the diffusion process. 
Although these approaches demonstrate proficiency in producing diverse and high-resolution textures, the iterative painting process frequently introduces 3D inconsistency and reduces efficiency.
Another issue is that while inpainting occurs iteratively, users lack the ability to interact with and precisely control the algorithm beyond providing a basic prompt. If the final results are not satisfactory, users are compelled to repaint the entire texture using a new random seed, precluding the possibility of partial editing of existing texture images.

In this study, we aim to tackle existing challenges by introducing a new Interactive text-to-texture framework, \textit{InteX}. 
We have created a user-friendly graphical interface that enables users to visualize and monitor the texture generation process, granting complete control over each step of the synthesis progress. 
This interface permits users to choose camera viewpoints for inpainting and to erase unwanted areas for further repainting. 
Additionally, we alleviate the issue of 3D consistency by training a unified depth-aware inpainting prior model on 3D datasets. 
This model is crucial for reducing 3D inconsistencies across multiple inpainting sessions. 
Moreover, our integrated diffusion process streamlines the entire generation workflow, significantly decreasing the texture generation time to approximately 30 seconds per instance. 
Compared to prior techniques, our method stands out for its enhanced controllability, efficiency, and flexibility in text-to-texture synthesis.

Our contributions can be summarized as follows:
\begin{itemize}
\item We introduce a novel text-to-texture framework that enables interactive visualization, inpainting, and partial repainting of texture, enhancing the flexibility and practicality of text-to-texture synthesis
\item We train a unified depth-aware inpainting diffusion prior on extensive 3D datasets to alleviate 3D inconsistency and boost generation speed by simplifying the pipeline.
\item Experimental results showcase the efficacy of the proposed method in generating high-quality textures coupled with smooth user interaction.
\end{itemize}

%% file: 2_related.tex
\subsection{2D Diffusion Models}
In recent years, notable strides in image generation have been witnessed, primarily attributed to the rapid evolution of diffusion models~\cite{ho2020denoising}. Through training on expansive text-image paired datasets~\cite{sharma2018conceptual, schuhmann2021laion}, diffusion models have demonstrated the capacity to learn an implicit correspondence between semantic concepts and textual prompts, thereby producing images that are both diverse and high-fidelity~\cite{radford2021learning, saharia2022photorealistic, nichol2022glide}. To mitigate the computational demands associated with large-scale datasets, Latent Diffusion Models (LDM)~\cite{rombach2022high} have been introduced, leveraging the diffusion model on the latent space instead of the pixel space.
Furthermore, several notable works~\cite{zhang2023adding, mou2023t2i, voynov2023sketch} have concentrated on integrating additional control signals such as sketches, edges, or depth maps. This adaptation of pre-trained diffusion models offers a controllable approach to the process.

\subsection{3D Content Generation}
Recent advancements in text-to-image diffusion models have also inspired substantial developments in the realm of 3D content generation. 
Initial endeavors employed CLIP~\cite{CLIP} to guide the optimization of 3D meshes~\cite{AvatarCLIP,CLIP-Mesh} or NeRF~\cite{DreamFields} based on textual prompts. However, the quality of the generated 3D models was often constrained, struggling to capture complex geometry or realistic textures.
The Dreamfusion approach~\cite{poole2022dreamfusion} introduces score distillation sampling (SDS) to harness the capabilities of powerful 2D diffusion models for 3D generation. Subsequent works~\cite{lin2023magic3d,michel2022text2mesh,wang2023prolificdreamer,chen2023fantasia3d,chen2023it3d,xu2023matlaber,huang2023dreamtime,metzer2022latent} have focused on enhancing geometry quality, appearance fidelity, and generation speed. 
For instance, camera-conditioned diffusion models~\cite{sweetdreamer,shi2023mvdream,zhao2023efficientdreamer} have been proposed to fortify geometry robustness. Additionally, 3D content generation can extend to accepting single-view images and performing image-to-3D transformations~\cite{liu2023zero,tang2023make,melas2023realfusion,xu2023neurallift,qian2023magic123}. Nevertheless, optimization-based 3D generation methods typically contend with slow generation speeds.
An alternative avenue of research focuses on inference-only 3D native models~\cite{nichol2022point,jun2023shap,hong2023lrm,liu2023one,gao2022get3d,wang2022rodin}. 
While these methods enable rapid 3D content generation in seconds, the quality of the generated content is often constrained. 
General 3D generation methods, capable of generating both geometry and texture from text prompts, face limitations in either generation speed or quality, particularly struggling to efficiently create high-quality textures with fixed geometry.

\subsection{Text-to-texture Synthesis}
Text-to-texture synthesis~\cite{metzer2022latent,richardson2023texture,yu2023texture,oechsle2019texture,siddiqui2022texturify,yu2021learning,chen2023text2tex,cao2023texfusion} has drawn significant inspiration from 2D diffusion models, yet with a distinctive focus on generating content on intricate 3D surfaces rather than on the image plane. 
Latent-NeRF~\cite{metzer2022latent} pioneers the application of SDS in a fixed-geometry context for texture generation. However, its generation speed is limited by the requirement for thousands of optimization steps, and the texture resolution is constrained.
TEXTure~\cite{richardson2023texture} explores an iterative depth-aware inpainting-based method, significantly improving generation speed. This method follows a predefined camera path to incrementally draw content on the mesh surface. 
Text2tex~\cite{chen2023text2tex} further advances this concept with an automatic view selection algorithm. 
While these approaches can synthesize high-quality textures, they often grapple with limited 3D consistency, attributed to the intricacies of depth-aware inpainting from multiple cameras.
Texfusion~\cite{cao2023texfusion} takes a different approach by interleaving the diffusion progress of multiple cameras to achieve intrinsic 3D consistency. However, it relies on a latent diffusion model~\cite{rombach2022high} and conducts texture diffusion in the latent space. 
An additional NeRF stage is needed to project the latent texture into the RGB space, adding complexity to the overall pipeline.
Our proposed method aligns with the iterative inpainting approach, but we focus on enhancing user interaction and simplifying the overall pipeline.

%% file: 3_method.tex
\begin{figure*}[t!]
    \centering
    \includegraphics[width=\textwidth]{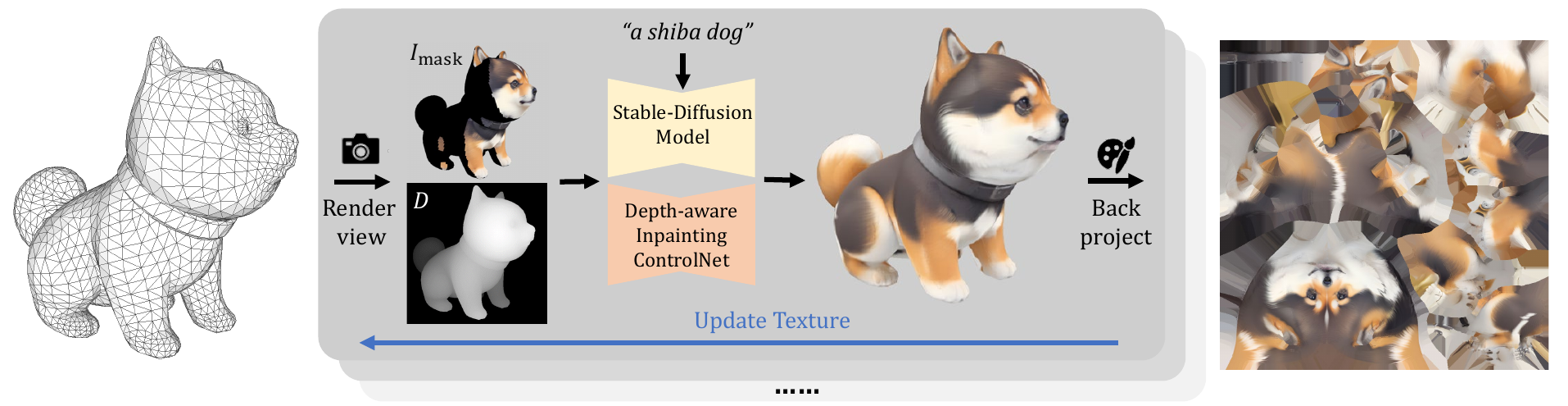}
    \caption{
    \textbf{Text-to-texture Synthesis Framework}.  We iteratively apply our unified depth-aware inpainting prior from multiple camera viewpoints and back-project the images to synthesize texture on the 3D surface.
    }
    \label{fig:framework}
\end{figure*}

In this section, we introduce our interactive texture synthesis framework as illustrated in Figure~\ref{fig:framework}.
Firstly, we train an unified depth-aware inpainting prior based on Stable-diffusion~\cite{rombach2022high} and ControlNet~\cite{zhang2023adding} (Section~\ref{sec:prior}).
Then, we propose an iterative texture synthesis algorithm using the 2D prior for efficient texture synthesis (Section~\ref{sec:texsyn}).
Lastly, we design a graphic user interface for interactive viewpoint selection and flexible repainting (Section~\ref{sec:gui}).

\subsection{Unified Depth-aware Inpainting Prior Model}
\label{sec:prior}

Unlike text-to-image synthesis, the process of text-to-texture synthesis involves the generation of content on 3D surfaces. 
This necessitates a comprehensive understanding of 3D information for the application of 2D diffusion priors. 
Previous methods~\cite{richardson2023texture,chen2023text2tex,cao2023texfusion} have attempted to tackle this issue by combining depth-to-image and image inpainting models~\cite{rombach2022high}.
However, these models are usually separately trained under distinct conditions, which constrains the quality of synthesized textures and leads to 3D inconsistency.
To overcome these challenges, we propose to \textbf{train a unified depth-aware inpainting prior model on 3D datasets}, which helps streamline the pipeline and concurrently enhance performance.

\subsubsection{Model Architecture.}
Our prior model aims to perform inpainting on a masked image guided by a given text prompt and additional depth information. 
We train the model based on the ControlNet architecture~\cite{zhang2023adding}, given its promising performance in both depth-to-image and image inpainting tasks.
The input to the model is a six-channel image. 
The first three channels represent the RGB values in $[0, 1]$, with the masked region designated by $-1$. 
The remaining three channels save the duplicated depth map, normalized to $[0, 1]$~\cite{zhang2023adding}. 
Control signals are injected into the denoising process of a pre-trained Stable-diffusion model~\cite{rombach2022high}. 
At denoising step $i$, the process is defined as follows:
\begin{equation}
\label{eq:denoise}
z_i = f_\psi(z_{i-1}, y, i, g_\phi(y, i, I_\text{mask} || D))
\end{equation}
where $z_i$ represents the latent image at timestep $i$, $y$ denotes the embeddings of the text prompt, $f_\psi$ signifies the noise prediction function with Stable-diffusion U-Net parameters $\psi$, and $g_\phi$ stands for the ControlNet function with parameters $\phi$. 
Additionally, $I_\text{mask}$ is the input image with the inpaint mask $M$, $D$ represents the depth image, and $||$ indicates concatenation.
To further enhance inpainting consistency, we incorporate latent mask blending~\cite{lugmayr2022repaint,avrahami2023blended}:
\begin{equation}
z_i \leftarrow z_i \odot m_i + z_{\text{mask},i} \odot (1 - m_i)
\end{equation}
where $m_i$ is the inpainting mask in the latent space, and $z_{\text{mask},i}$ is the latent code of $I_\text{mask}$ with injected noise at timestep $i$.

\subsubsection{Model Learning.}
\begin{figure*}[t!]
    \centering
    \includegraphics[width=\textwidth]{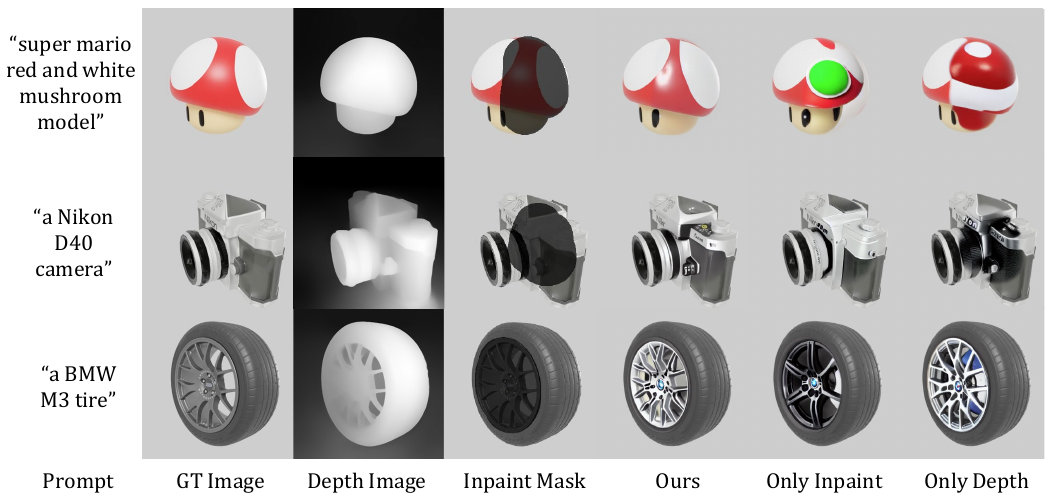}
    \caption{
    \textbf{Depth-aware inpainting results of the 2D prior}. 
    Our method produces best depth-aligned inpainting results, while inpainting-only may change the geometry and depth-only may produce inconsistent content.
    }
    \label{fig:comp2d}
\end{figure*}

To achieve the primary objective of synthesizing textures on common 3D shapes, our model is trained on the Objaverse~\cite{deitke2023objaverse} dataset, using both text descriptions and rendered images from Cap3D~\cite{luo2023scalable}. 
Similar to the approach adopted by other methods~\cite{shi2023mvdream}, we perform dataset filtering, retaining a subset featuring high-quality textures for the training process.
Due to resource constraints, we opt to estimate the depth map for each rendered image using DPT~\cite{Ranftl2021, Ranftl2020} rather than rendering the depth from the original 3D models. Our experiments indicate that the estimated depth maps are sufficiently accurate in most cases, as illustrated in Figure~\ref{fig:comp2d}.
During training, inpainting masks are generated dynamically. 
We employ a hybrid mask generation strategy to better simulate practical inpainting scenarios:
(1) The entire object is masked.
(2) Either the left or right half of the object is masked.
(3) The object's overlap with a randomly positioned ellipse or rectangle is masked.
(4) Regions with a depth value below a randomly determined threshold are masked.
Our base model is Stable-diffusion v1.5~\cite{rombach2022high}, and we initialize the ControlNet parameters with the official v1.1 inpainting model~\footnote{\url{https://github.com/lllyasviel/ControlNet-v1-1-nightly}}, with the exception of the different first layer. 
Only the ControlNet parameters undergo training, enabling the resulting model to be applied to various personalized Stable-diffusion checkpoints.

\subsection{Iterative Texture Synthesis}
\label{sec:texsyn}

Leveraging the depth-aware inpainting prior model alongside a 3D shape $M$, we can perform text-to-texture synthesis as outlined in Figure~\ref{fig:framework}. 
Our approach adopts the iterative inpainting methodology~\cite{richardson2023texture, chen2023text2tex} for synthesizing textures on 3D surfaces. 
The key difference from previous methods is that \textbf{our depth-aware inpainting prior model facilitates a simpler and more efficient pipeline, eliminating the necessity for any optimization~\cite{richardson2023texture} or multi-stage refinement processes~\cite{chen2023text2tex}}, while still achieving comparable performance.
To begin with, the 3D shape is initialized with an empty texture image $ T \in \mathbb{R}^{H \times W \times 3} $, a weight image $ W \in \mathbb{R}^{H \times W} $, and a view cosine cache image $ V \in \mathbb{R}^{H \times W} $. 
We then sequentially process all camera viewpoints, executing \textit{rendering}, \textit{inpainting}, and \textit{updating} at each step.

\subsubsection{Rendering.}
Firstly, we render the 3D shape with the current texture image to obtain the necessary image buffers, including an RGB image with inpainting mask $ I_{\text{mask}} $, a depth map $ D $, a normal map $ N $, and UV coordinates $ U $. 
Using the normal map, we compute the cosine value of the view direction (from surface to camera) and the surface normal at each pixel. 
Following~\cite{richardson2023texture,chen2023text2tex}, we leverage this cosine map and the weight image to partition the image into a trimap denoted as $ M_{\text{generate}} $, $ M_{\text{refine}} $, and $ M_{\text{keep}} $. 
Specifically, $ M_{\text{generate}} $ encompasses the region viewed for the first time with $ W = 0 $, $ M_{\text{refine}} $ encompasses the seen region with a smaller view cosine cache compared to the current view, and $ M_{\text{keep}} $ covers the remainder of the seen region.

\subsubsection{Inpainting.}
Subsequently, the pretrained prior model is used to generate an inpainted output image $I$ in Equation~\ref{eq:denoise}. 
Regarding the latent blending mask $m$, it is expanded from the \textit{generate} region to include the \textit{refine} region during the denoising process. 
This expansion allows for the appropriate alteration of the \textit{refine} region:
\begin{equation}
m_i =
\begin{cases}
M_\text{generate}, & i \le (1 - \alpha) \times L \\
M_\text{generate} \cup M_\text{refine}, & i > (1 - \alpha) \times L \\
\end{cases}
\end{equation}
where $L$ denotes the total number of denoising steps, and $\alpha$ is the refining strength.
The \textit{keep} region remains strictly fixed through mask blending in the image space:
\begin{equation}
I \leftarrow I \odot (1 - M_\text{keep}) + I_\text{mask} \odot M_\text{keep}
\end{equation}
Since the output resolution of Stable-diffusion v1.5 is constrained to $512 \times 512$, a super-resolution step~\cite{wang2021realesrgan} is performed on the output image $I$ to allow higher-resolution texture.

\subsubsection{Updating.}
\begin{figure}[t!]
    \centering
    \includegraphics[width=0.9\linewidth]{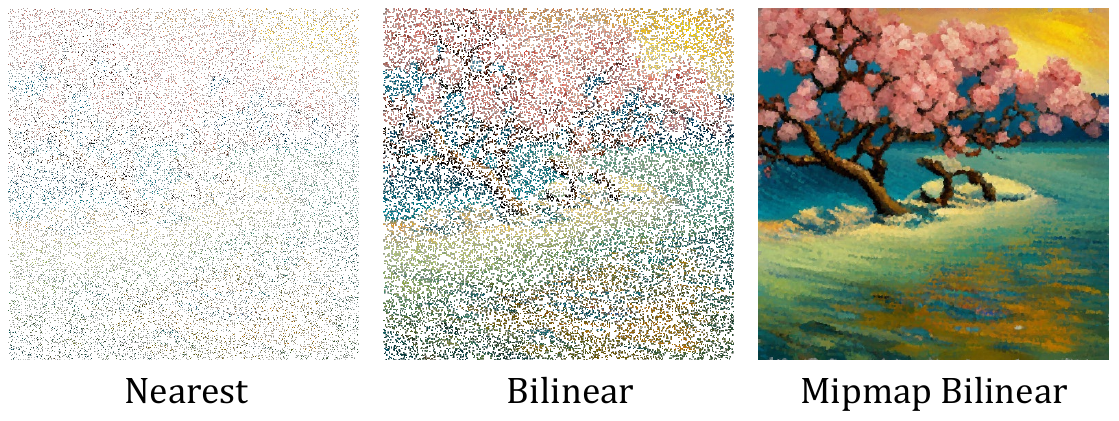}
    \caption{
    \textbf{Mipmap Bilinear Extrapolation}. We randomly sample 10\% pixels and put them back to an empty image using different extrapolation methods. 
    The proposed mipmap bilinear method can fill most of the holes.
    }
    \label{fig:grid_put}
\end{figure}

Finally, the texture space buffers are updated by back-projecting the generated image $I$ through UV mapping. However, mapping pixels from a low-resolution image $I$ to a high-resolution image $T$ may result in holes, as depicted in Figure~\ref{fig:grid_put}. To address this, we propose a mipmap bilinear extrapolation method for anti-aliasing. The target texture image $T$ is initially down-scaled to form a multi-scale pyramid. The algorithm then operates as the inversion of bilinear interpolation, reintegrating each pixel from $I$ at each mipmap level. Additionally, the pixel weights are updated to $W$, and the view cosine cache $V$ is updated to keep the maximum value from all seen cameras.

Upon completion of all camera viewpoints, a simple post-processing step is applied by dilating the texture map~\cite{stable-dreamfusion} and incorporating super-resolution~\cite{wang2021realesrgan}. This step enhances the sharpness of the texture and mitigates visible seams resulting from UV unwrapping.

\subsection{GUI for Practical Use}
\label{sec:gui}
Previous iterative texture synthesis methods have two limitations that hinder practical use:
(1) They require the selection of an appropriate set of camera viewpoints. 
While it is possible to use a pre-defined set of camera viewpoints for automatic inpainting~\cite{cao2023texfusion,richardson2023texture}, this approach may be sub-optimal for diverse 3D shapes, particularly when the shapes are not axis-aligned with the camera coordinate system. 
Text2tex~\cite{chen2023text2tex} introduces an automatic camera selection algorithm for refinement, yet it still relies on a preset for initialization.
(2) Another limitation is the inflexibility and opacity of the generation progress. 
For instance, the iterative inpainting process may yield undesired results on a specific region of the texture, while the other regions are satisfactory. 
Users might wish to only \textit{repaint} this specific region while keeping the other regions unchanged. 
However, previous methods can only regenerate the entire texture with a different random seed.

To address these challenges and enhance practical utility, we propose the development of a Graphic User Interface (GUI) based on our algorithm. 
As depicted in Figure~\ref{fig:teaser}, the GUI functions as a 3D viewer, allowing the orbiting of the camera around the object for inpainting. 
It also provides the capability to erase part of the texture by back-projecting a user-defined 2D mask onto the 3D surface. 
This way, only the selected texture region can be erased and repainted. 
We further allow changing the text prompts at any time of generation for different details.
This approach enhances the controllability and flexibility of our synthesis process, enabling users to scrutinize inpainting results at each step and make informed decisions on whether to repaint specific regions.

%% file: 4_exps.tex
\subsection{Implementation Details}

\subsubsection{Dataset and Training.}
We use a simple strategy to filter out meshes without texture (mostly white if rendered) in the Objaverse~\cite{deitke2023objaverse} dataset since they are not helpful for training our texture synthesis model.
we discard the object if the minimal pixel value of its rendered RGB image is bigger than $0.5$.
This leaves us $550$k objects for training, and we use $8$ rendered images per each object from Cap3D~\cite{luo2023scalable}.
The ControlNet training is executed on $4$ NVIDIA V$100$ ($32$G) GPUs, employing a batch size of $4$ per GPU. 
Gradients are accumulated to achieve a valid batch size of $32$ before back-propagation. 
The learning rate is held constant at $5\times 10^{-6}$. Training the model on the filtered dataset takes approximately 5 days for 10 epochs.

\subsubsection{Inference.}
For the text-to-texture synthesis algorithm, it requires less than $8$GB GPU memory and takes around $30$ seconds to generate with an NVIDIA V$100$ ($16$GB) GPU.
We append extra positive prompts of \textit{``masterpiece, high quality''} and negative prompts of \textit{``bad quality, worst quality, shadows''} by default.
Directional prompts of \textit{``front/side/back view''} is appended if the object can be oriented following~\cite{poole2022dreamfusion,richardson2023texture}. 
The camera radius is set to $2.5$ with a field of view (FOV) of $50$ degree along the Y-axis.
The resolution of the texture is set to $1024 \times 1024$, and the inpainting image is also set to $1024 \times 1024$ by super-resolution~\cite{wang2021realesrgan}.
We choose the UniPC~\cite{zhao2023unipc} sampler for the diffusion model, and use $20$ steps for each inference.
The refining strength $\alpha$ is set to $0.4$.
Ten pre-defined camera viewpoints are used for the automatic mode, with each view taking about $3$ seconds to finish inpainting, leading to a total time of $30$ seconds to generate a whole texture.

\subsection{Effectiveness of Depth-aware Inpainting}
We first perform experiments to verify the effectiveness of our depth-aware inpainting prior model.
In Figure~\ref{fig:comp2d}, we compare our depth-aware inpainting results to the baseline inpainting-only results.
We show that our model can correctly handle both inpainting and depth information, producing image with both depth and content aligned to the input.
Only using ControlNet for inpainting leads to an altered geometry, which is prohibitive in text-to-texture synthesis.
On the other hand, only using ControlNet for depth leads to inconsistent texture even if we apply mask blending~\cite{avrahami2023blended,lugmayr2022repaint}, which is the major source of 3D inconsistency in previous works.

\begin{figure*}[ht!]
    \centering
    \includegraphics[width=0.97\textwidth]{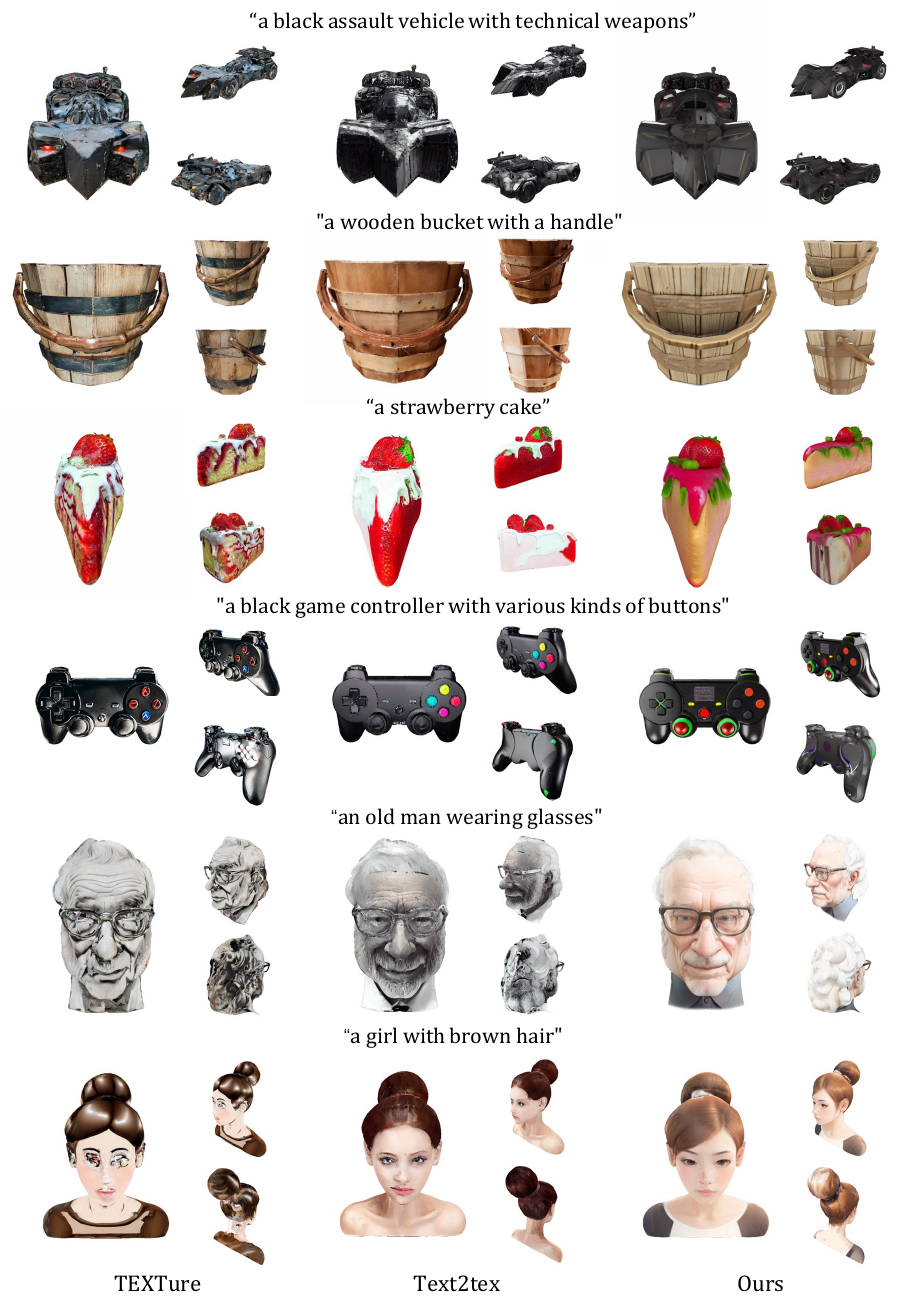}
    \caption{
    \textbf{Qualitative comparison with recent methods}. Our method generates textures with higher quality and better 3D consistency.
    }
    \label{fig:comp}
\end{figure*}

\begin{figure}[t!]
    \centering
    \includegraphics[width=\linewidth]{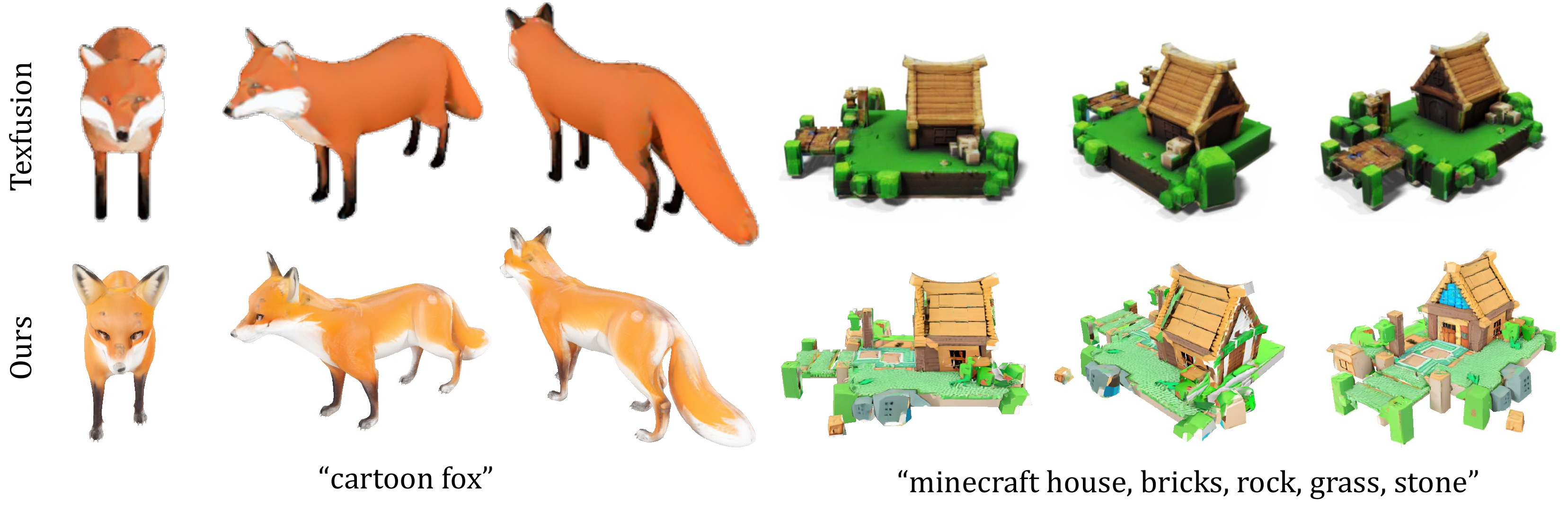}
    \caption{
    \textbf{Qualitative comparison} with Texfusion~\cite{cao2023texfusion}. We use the closest mesh we can find to run our method.
    }
    \label{fig:comptexf}
\end{figure}

\begin{figure*}[t!]
    \centering
    \includegraphics[width=\textwidth]{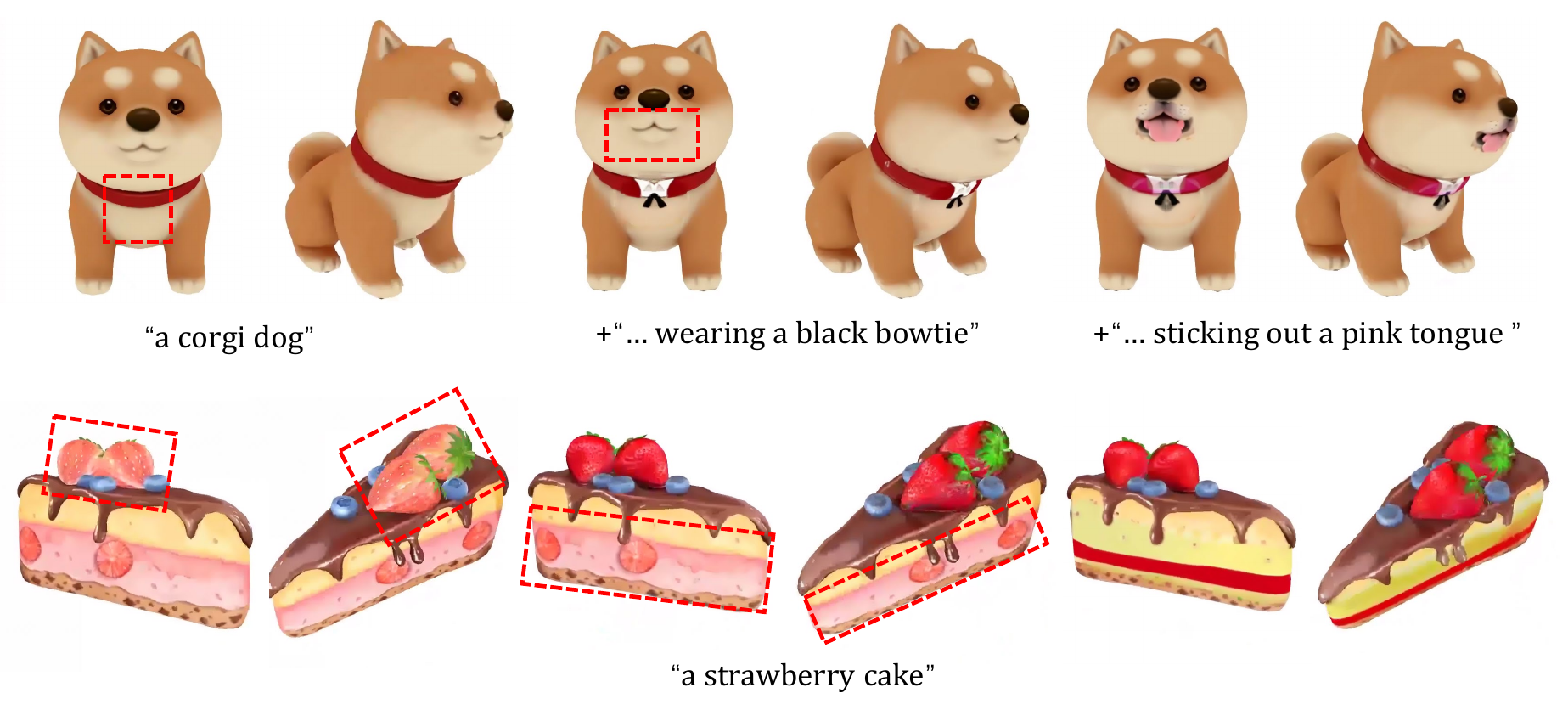}
    \caption{
    \textbf{Results of text-guided sequential editing}. 
    Our method supports re-painting texture of selected regions guided by different text prompts, or simply generating different patterns under the same prompt. The red dashed boxes refer to the selected areas for repainting.
    }
    \label{fig:redraw}
\end{figure*}

\begin{figure*}[t!]
    \centering
    \includegraphics[width=\textwidth]{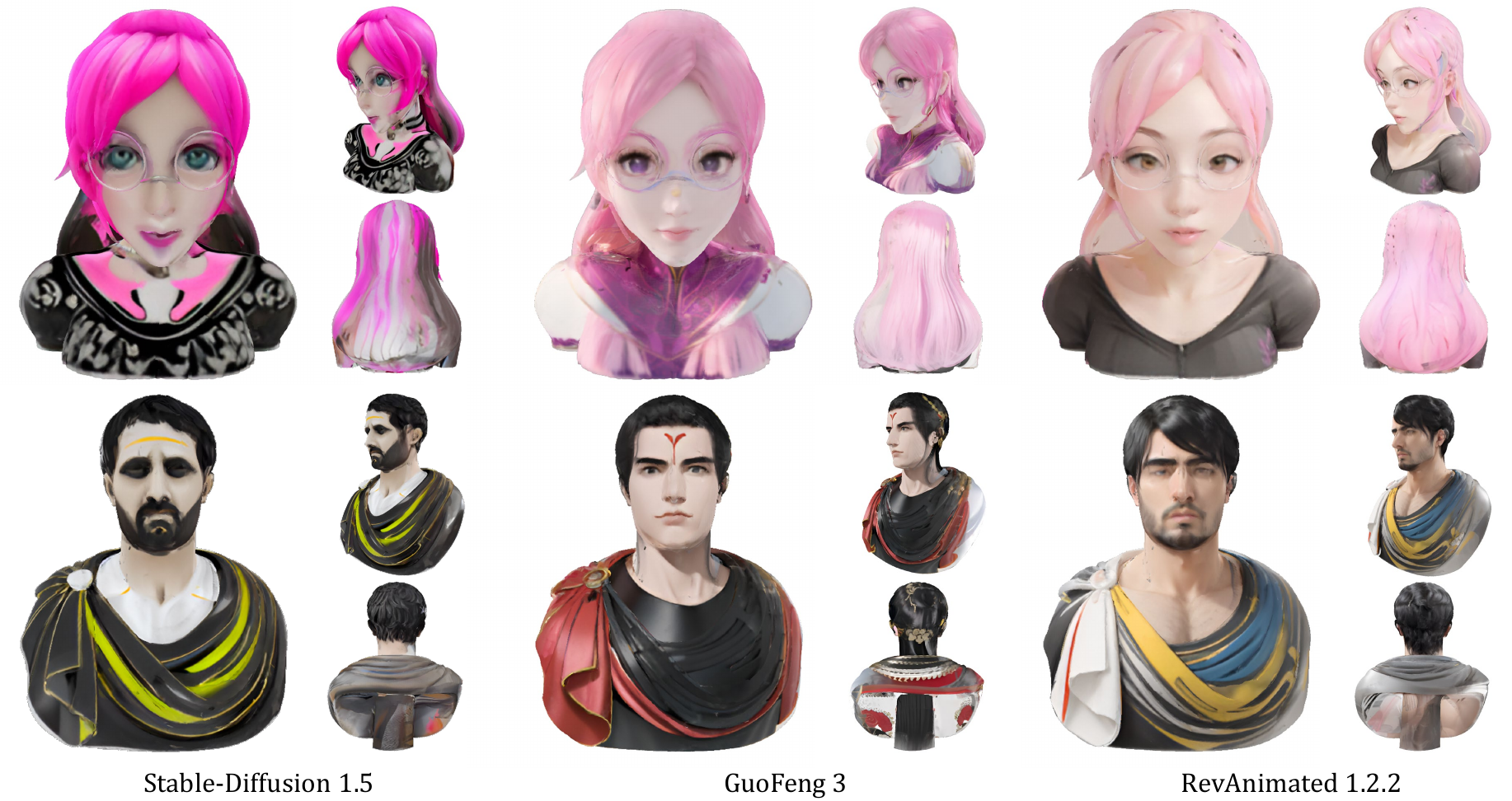}
    \caption{
    \textbf{Results of different Stable-diffusion base models}. 
    The prompts are ``a girl with pink hair'' and ``a man with black hair''.
    }
    \label{fig:custom_sd}
\end{figure*}

\begin{figure*}[t!]
    \centering
    \includegraphics[width=\textwidth]{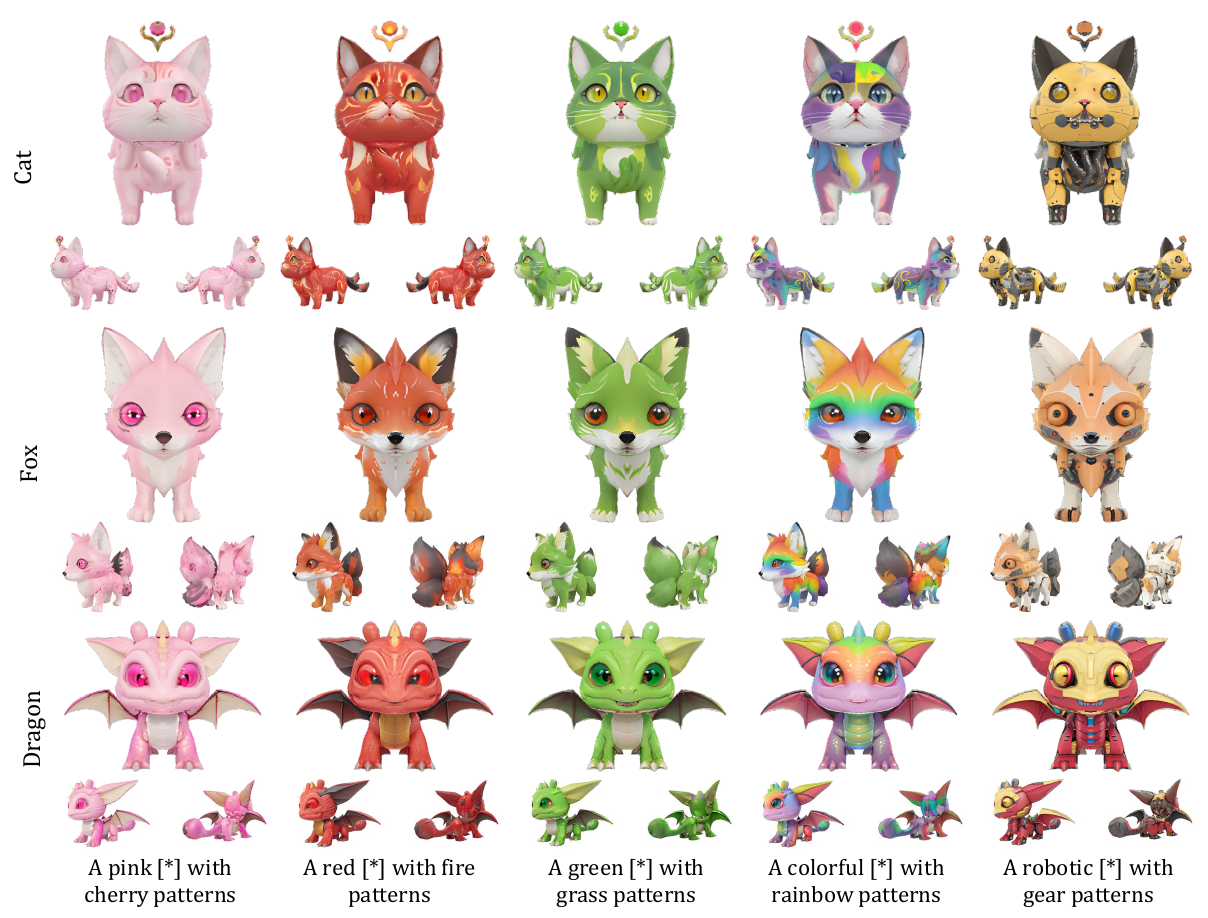}
    \caption{
    \textbf{Results of different prompts on the same geometry}. 
    We show generating themed skins for animal mesh assets from games.
    }
    \label{fig:pet}
\end{figure*}

\subsection{Comparisons with State-of-the-Arts}

We then apply our 2D prior model on text-to-texture synthesis experiments, and compare against recent methods to show the advantages of our method.

\subsubsection{Qualitative Comparisons.}

In Figure~\ref{fig:comp}, we present visualizations of multi-view rendering for synthesized textures generated by our method, compared against recent works~\cite{richardson2023texture,chen2023text2tex}. Our technique demonstrates superior performance in generating textures with enhanced detail and improved 3D consistency.
In Figure~\ref{fig:comptexf}, we compare against Texfusion~\cite{cao2023texfusion}. Since their method is not open-sourced and some used data are not available, we use the results from their paper, and run ours with the closest data we could found to compare.
Further elaborated in Figure~\ref{fig:redraw}, our design facilitates the re-painting of selected regions via a graphical user interface. This feature supports sequential editing, enabling the incremental enhancement of texture quality. Our approach streamlines the modification process, offers precise control over texture improvements, and guarantees heightened fidelity and detail in the final textures.
Another advantage of our method is its independence from any specific fine-tuned Stable-diffusion model. This permits the application of various personalized checkpoints to create textures in multiple styles, as exemplified in Figure~\ref{fig:custom_sd}. By employing the same prompt with different Stable-diffusion checkpoints (GuoFeng 3~\footnote{\url{https://huggingface.co/xiaolxl/GuoFeng3}} and RevAnimated 1.2.2~\footnote{\url{https://civitai.com/models/7371/rev-animated}}), our method yields a rich diversity of stylistic outputs.
Moreover, Figure~\ref{fig:pet} illustrates the capacity of our model to produce a wide range of styles through variations in prompts, further demonstrating the method's adaptability and broad applicability in generating customized texture styles.

\subsubsection{Quantitative Comparisons.}
In Table~\ref{tab:userstudy}, we compare the generation time and perform an user study to evaluate the 3D consistency and overall texture quality against recent works~\cite{richardson2023texture,chen2023text2tex}.
We render 360 degree rotating videos of 3D meshes with generated textures from different methods.
There are in total 45 videos for 3 methods (TEXTure~\cite{richardson2023texture}, Text2tex~\cite{chen2023text2tex}, and our method) to evaluate.
Each volunteer is shown 15 samples containing the input prompt and a rendered video from a random method, and asked to rate in two aspects: 3D consistency and overall texture quality. We collect results from 30 volunteers and get 450 valid scores in total.
Our method achieves $10$ times acceleration over previous iterative inpainting-based methods using 10 cameras, while also producing more preferred texture quality with a user-friendly GUI.

\begin{figure*}[t!]
    \centering
    \includegraphics[width=\textwidth]{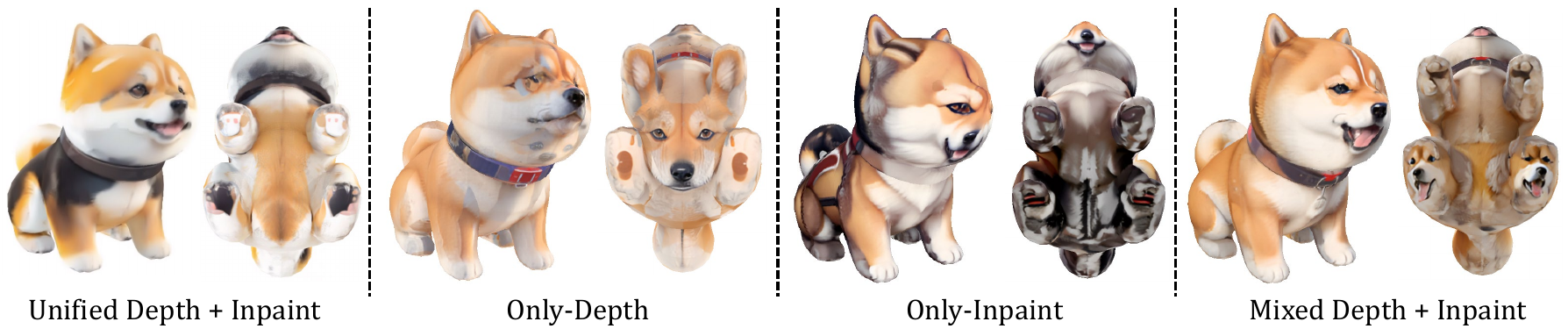}
    \caption{
    \textbf{Ablation study} on the effect of different 2D diffusion prior models.
    }
    \label{fig:ablation}
\end{figure*}

\begin{figure}[t!]
    \centering
    \includegraphics[width=\linewidth]{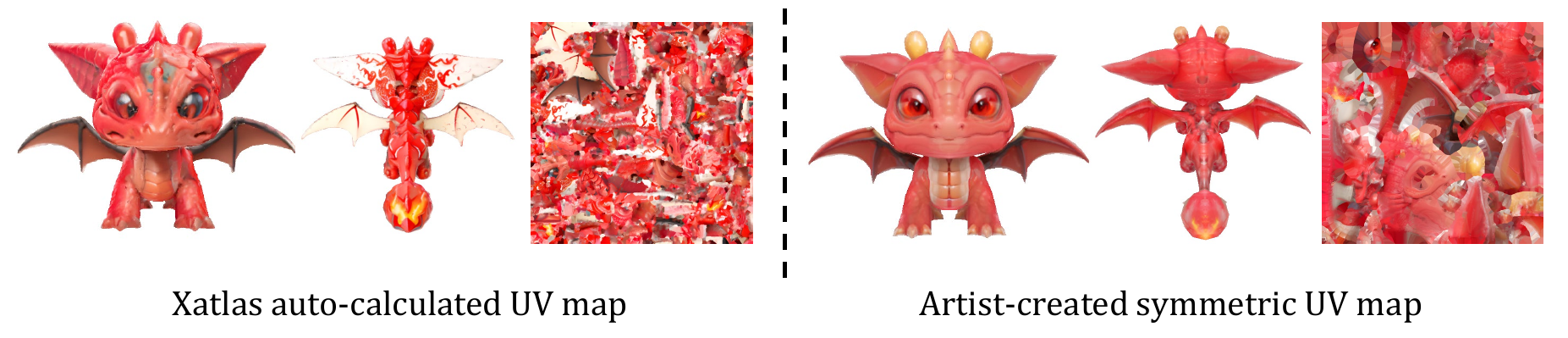}
    \caption{\textbf{Ablation study} on auto-generated \textit{v.s.} artist-created UV maps.}
    \label{fig:uv}
\end{figure}

\input{tabs/user_study}

\subsection{Ablation Study}

\subsubsection{Diffusion Priors.}
We first carry out ablation studies on the 2D diffusion prior models.
In Figure~\ref{fig:ablation}, we compare against other choices of prior models.
Similar to the observation in 2D experiments, depth-only model cannot generate consistent texture and fail to use the already inpainted content.
Inpaint-only model fails to preserve the original depth and leads to deformed result when rendering from different views.
We also try mixing multiple priors similar to TEXTure~\cite{richardson2023texture}, by interleaving the denoising process of a depth-only model and an inpaint-only model. However, it still tends to generate inconsistent information from difficult views due to the separately trained depth-to-image and inpainting models.

\subsubsection{UV Coordinates.}
In Figure~\ref{fig:uv}, we also compare the texture synthesis results on artist-created UV maps with auto-generated UV maps.
In the case of artist-created symmetric meshes, it is typical to reuse the texture map (\textit{e.g.}, both eyes can share the same texture map area). 
Like prior methods, we prefer the original UV map from the mesh when available, which maintains symmetry in generated texture. 
Alternatively, discarding the original UV map and unwrapping it based on geometry is also feasible, where our method still generates satisfactory texture.

\subsection{Limitations}
Despite efforts to enhance the practicality of texture synthesis, our method still encounters limitations. 
One major drawback is the single-view rendering's failure to encompass all dimensions of the 3D model, which may result in 3D inconsistencies during the iterative inpainting progress, especially when using unsuitable camera choices.
Nevertheless, with recent advancements in multi-view diffusion models~\cite{shi2023mvdream,liu2023syncdreamer}, we expect that developing a multi-view, depth-aware inpainting model could successfully address the issue of 3D inconsistency.

%% file: tabs/user_study.tex
\begin{table}[t!]
\begin{center}
\begin{tabular}{lccc}
\hline
Method & 3D Consistency ($\uparrow$) & Overall Quality ($\uparrow$)  & Inference Time ($\downarrow$) \\
\hline
TEXTure~\cite{richardson2023texture} & 3.57 & 2.89 & 5 min \\ 
Text2tex~\cite{chen2023text2tex}     & 3.53 & 3.06 & 15 min \\ 
Ours                                 & \textbf{4.24} & \textbf{3.81} & \textbf{30 sec} \\ 
\hline
\end{tabular}

\end{center}
\caption{
\textbf{User Study and Inference Time}. The rating is of scale 1-5, the higher the better.
}


\label{tab:userstudy}
\end{table}

%% file: 5_conclusion.tex
In this paper, we present \textit{InteX}, an interactive text-to-texture framework designed to improve the practical aspects of the texture synthesis task. 
Utilizing a specialized graphical user interface (GUI), users are provided with unparalleled control over the texture synthesis process, facilitating more accurate and preferred results. 
Our approach also alleviates the challenges of 3D consistency and enhances generation speed in text-to-texture synthesis through the adoption of a unified depth-aware inpainting diffusion prior model. 
Extensive experiments illustrate the effectiveness and efficiency of our method.